\documentclass[runningheads]{llncs}

\usepackage[T1]{fontenc}
\usepackage[utf8]{inputenc}

\usepackage[square, comma, numbers, sort&compress, sectionbib]{natbib}
\renewcommand{\refname}{References}

\usepackage{amsmath, amssymb, bm}
\usepackage{microtype}

\usepackage[table]{xcolor}
\usepackage{graphicx, verbatim}

\usepackage{booktabs}
\usepackage{multirow}
\usepackage{makecell}
\usepackage{colortbl}
\usepackage{arydshln}
\usepackage{threeparttable}
\usepackage{adjustbox}

\newcommand{\modelbase}{\textsc{Mr}}
\newcommand{\modelname}{\modelbase.\textsc{Dec}}
\newcommand{\github}{https://github.com/yejix-ai/MR.DEC}

\begin{document}

\title{\modelname{}: Daily-Scale Longitudinal Multimodal Modeling for 30-Day Readmission Prediction}
\titlerunning{\modelname{}} 

\author{Minjun Kim\thanks{Equal contribution}\orcidID{0009-0009-8429-5240} \and 
Jong Hak Moon\textsuperscript{*}\orcidID{0000-0002-6708-3918}}

\index{Kim, Minjun}
\index{Moon, Jong Hak}

\authorrunning{Kim et al.}

\institute{Yeji X, Seoul, South Korea\\
\email{\{mj.kim, jh.moon\}@yejix.ai}}

\maketitle

\begin{abstract}
Predicting 30-day hospital readmission is essential for assessing patient stability and optimizing healthcare resources. As clinical risk evolves with the accumulation of evidence during hospitalization, capturing these dynamic trajectories is essential. 
However, many existing approaches compress the complex longitudinal history into fixed representations, often losing the granular, day-level clinical signals that reflect a
patient's evolving physiological state. 
To address this, we propose \modelname{} (\textbf{M}ultimodal \textbf{R}eadmission-risk prediction \textbf{De}coder), which models each admission as a natural chronological sequence of daily multimodal events. By leveraging a Transformer Decoder, \modelname{} integrates daily Electronic Health Record(EHR) updates and intermittent Chest X-ray(CXR) findings in a time-aligned stream, reflecting the actual clinical workflow. To ensure robustness, we utilize Disease-Specific Supervised Contrastive Learning as an auxiliary regularization to induce a diagnosis-aware structure in the latent space. Evaluations on the MIMIC-IV and MIMIC-CXR datasets show that \modelname{} achieves state-of-the-art performance by preserving the integrity of the clinical sequence. Furthermore, our model identifies "Critical Days" within an admission, providing actionable and clinically grounded interpretations for real-time risk stratification. Code is available at: \github{}


\end{abstract}

\section{Introduction}
\begin{figure}[t]
  \centering
\includegraphics[width=0.8\linewidth]{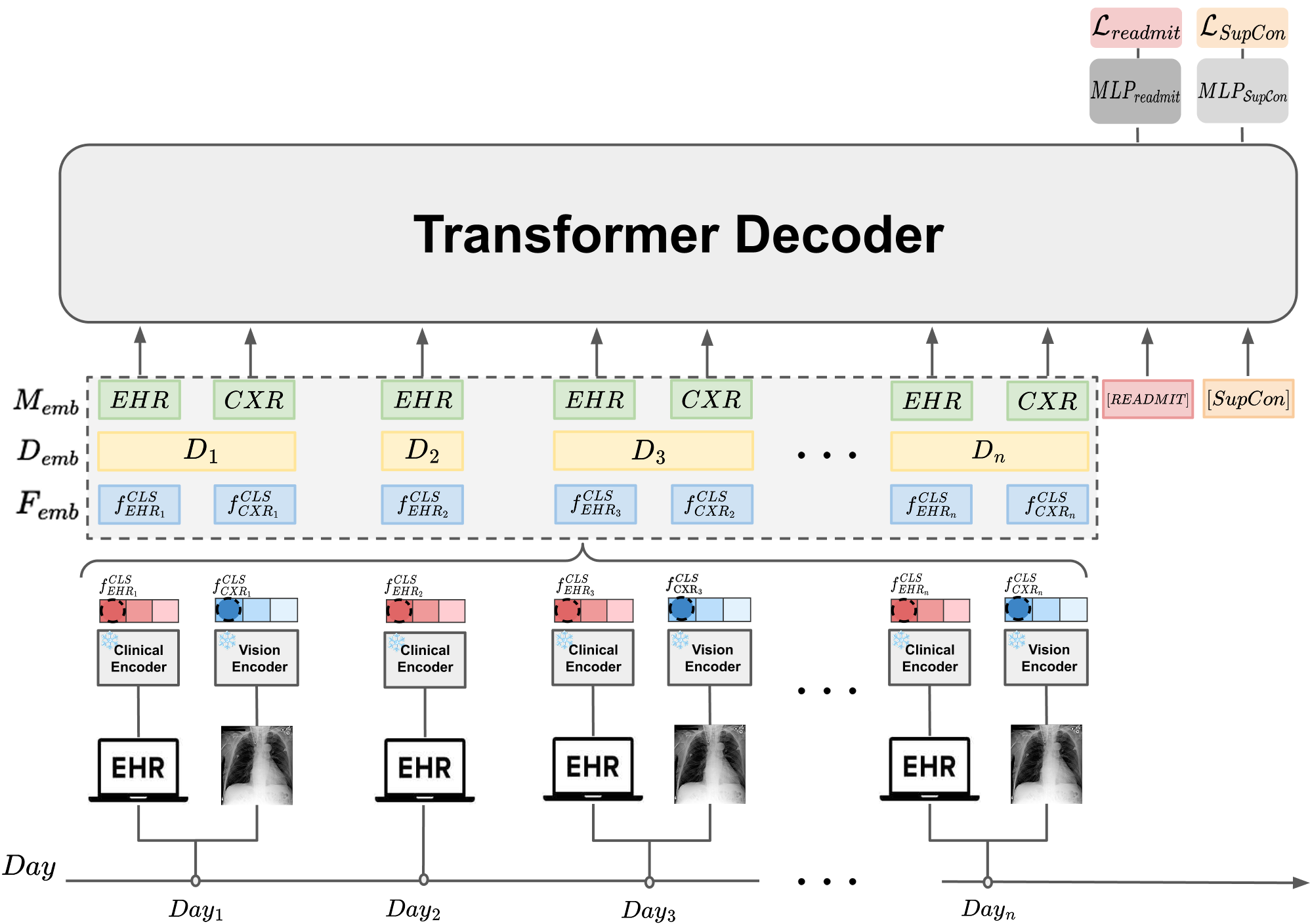}  
\caption{\footnotesize \textbf{\modelname{} (Multimodal Readmission-risk Prediction Decoder).}
For each hospital day $n$, textualized EHR (e.g., ICD codes mapped to natural-language descriptors) is encoded by a clinical encoder and same-day CXR is encoded by a vision encoder, producing modality-specific \texttt{[CLS]} representations $\{f_{\mathrm{EHR}_n}^{\mathrm{CLS}},\, f_{\mathrm{CXR}_n}^{\mathrm{CLS}}\}$ (with $f_{\mathrm{CXR}_n}^{\mathrm{CLS}}$ omitted when no CXR is acquired).
Day-wise features are serialized in chronological order and projected with modality-specific layers; modality and day embeddings are added to obtain the embedded token sequence $F_{\mathrm{emb}}=(h_1,\ldots,h_L)$.
We then append two learnable special tokens, yielding the decoder input $\tilde{F}_{\mathrm{emb}}=[F_{\mathrm{emb}}, \texttt{[READMIT]}, \texttt{[SupCon]}]$.
The final hidden states of \texttt{[READMIT]} and \texttt{[SupCon]} are fed to separate MLP heads for readmission risk estimation optimized with $\mathcal{L}_{\mathrm{readmit}}$ and admission-level representation learning optimized with $\mathcal{L}_{\mathrm{SupCon}}$, respectively.}
  \label{fig:model}
\end{figure}

Predicting 30-day all-cause hospital readmission is a widely used benchmark of healthcare quality and patient stability. Unplanned readmissions worsen outcomes and impose substantial economic burden, costing billions annually. While post-discharge environmental factors also shape risk \cite{social1, social2, social3}, a central engineering objective is to maximize discriminative signal from in-hospital data to identify high-risk cohorts before discharge \cite{alarcon2024prediction, diabetic_readmit, must, nvidia, readmission, readmission2, stgnn}. The operational goal is a robust screening mechanism that interprets a patient’s evolving trajectory to support triage and resource allocation. Accurate longitudinal modeling is essential to capture subtle signs of instability that may warrant intervention.

Toward this end, readmission modeling has evolved toward richer patient representations. Early approaches relied on administrative variables such as ICD codes as static categorical features \cite{diabetic_readmit, okolie2025machine}, but these lack the granularity needed to reflect complex clinical progression. More recent work adopted multimodal learning, recognizing that structured codes alone are insufficient. This includes incorporating clinical notes with LLMs \cite{nvidia, must} and integrating CXR using deep neural networks \cite{stgnn, must, readmission, readmission2, diabetic_readmit}. Such fusion improves prediction by combining complementary signals, including semantic context from text and visual evidence from imaging.

Within this multimodal landscape, fusing time-series EHR and routine CXR provides a strong foundation for day-level trajectory tracking. Although clinical notes can add semantic detail, they are often irregular, subjective, and retrospective \cite{nvidia, must}, which can introduce reporting delays and observer bias. In contrast, time-series EHR (e.g., labs) and CXR serve as more direct proxies of physiological status: numerical signals provide immediate quantitative indicators, while CXR offers an observer-independent view of internal pathology such as pulmonary edema progression that may not be captured by codes \cite{stgnn}. High-fidelity tracking of daily clinical evolution therefore benefits from emphasizing these objective snapshots. However, existing SOTA (state-of-the-art) systems such as MM-STGNN \cite{stgnn} and MuST \cite{must} often aggregate variable-length admissions into static representations to exploit population-level structure, which can dilute intra-patient temporal granularity and weaken causal dependencies. This motivates an alternative design that prioritizes sequence fidelity over global graph topology, especially when the timing of physiological change is clinically decisive.

To bridge this gap, we propose \modelname{} (\textbf{Multimodal Readmission-Risk Prediction Decoder}), a daily-scale causal Transformer that preserves intra-admission temporal structure by modeling each hospitalization as a chronological token stream of daily EHR records interleaved with sporadic same-day CXR observations. \textbf{Our contributions are threefold:} (i) we introduce a new multimodal trajectory architecture that jointly models longitudinal EHR and intermittent CXR and achieves SOTA performance on 30-day readmission prediction, (ii) we propose an efficient daily-recording tokenization that supports longer sequences for fine-grained risk tracking across the full hospital stay without retraining separate unimodal models, and (iii) we develop clinically grounded training and analysis components, including disease-specific supervised contrastive regularization for robustness under label noise and imbalance, and gradient-based "Critical Days" identification that links
predicted risk changes to day-level EHR and CXR.

\section{Methods}
\subsubsection*{Problem Formulation}
\label{sec:problem}

We formulate 30-day all-cause readmission prediction as a \textit{causal in-hospital trajectory modeling} task, motivated by evidence that readmission risk evolves as clinical evidence accumulates during hospitalization \citep{jiang2019readmission, davis2024survival}.
A single admission is represented as a chronological sequence of hospital days
$\mathcal{S}=(d_1,d_2,\ldots,d_N)$ over $N$ days, where each day is a multimodal record
$d_n := \big(x_n^{\text{EHR}}, x_n^{\text{CXR}}\big)$.
Here, $x_n^{\text{EHR}}$ denotes the structured EHR observed on day $n$, and
$x_n^{\text{CXR}}$ denotes the chest radiograph acquired on day $n$ (if available; otherwise $x_n^{\text{CXR}}=\varnothing$).
Given the full in-hospital history $\mathcal{H}_N=(d_1,\ldots,d_N)$, our model predicts the admission-level outcome
\[
\hat{y} = f_{\theta}(\mathcal{H}_N),
\]
while enforcing causality inside the decoder via masked self-attention so that each token aggregates evidence only from preceding tokens in the interleaved stream.

\subsection{\modelname{} (\textbf{M}ultimodal \textbf{R}eadmission-risk prediction \textbf{De}coder)}
\subsubsection*{Hierarchical Multimodal Tokenization}
\label{sec:tokenization}

We encode the full admission history $\mathcal{H}_N$ as a day-aligned, variable-length token stream that reflects the fact that CXR is not acquired every day (Fig.~\ref{fig:model}). For each hospital day $n \in \{1,\ldots,N\}$, textualized EHR is encoded by a frozen Clinical Encoder (e.g., BioClinical ModernBERT~\citep{modernbert}) to obtain $f_{\text{EHR}_n}^{\text{CLS}}$, and same-day CXR (if present) is encoded by a frozen Vision Encoder (e.g., EVA-X-Base~\citep{eva-x}) to obtain $f_{\text{CXR}_n}^{\text{CLS}}$.

We serialize day-wise tokens in chronological order by emitting one EHR token per day and inserting the CXR token immediately after it only on imaging days.
We denote the resulting interleaved token features for the admission as
\[
F = (f_1,\ldots,f_L),
\]
where
\[
{\scriptsize
F
=
\Big[
\underbrace{f_{\text{EHR}_1}^{\text{CLS}},\, f_{\text{CXR}_1}^{\text{CLS}}}_{\text{Day }1},
\underbrace{f_{\text{EHR}_2}^{\text{CLS}}}_{\text{Day }2},
\underbrace{f_{\text{EHR}_3}^{\text{CLS}},\, f_{\text{CXR}_3}^{\text{CLS}}}_{\text{Day }3},
\dots,
\underbrace{f_{\text{EHR}_N}^{\text{CLS}},\, f_{\text{CXR}_N}^{\text{CLS}}}_{\text{Day }N}
\Big].
}
\]
Thus, the token length is
\[
L = N + \sum_{n=1}^{N} \mathbb{I}\!\left[x_n^{\text{CXR}}\neq\varnothing\right].
\]

Each token $f_i$ has modality $m_i \in \{\text{ehr},\text{cxr}\}$ and hospital-day index $n_i \in \{1,\ldots,N\}$.
We project $f_i$ into the model dimension $d_{\text{model}}$ with modality-specific projections and add modality and day embeddings:
\begin{equation}
h_i
=
\text{GELU}\!\big(W_{\text{proj}}^{m_i} f_i + b_{\text{proj}}^{m_i}\big)
+ M_{\text{emb}}(m_i)
+ D_{\text{emb}}(n_i),
\end{equation}
where $M_{\text{emb}}(\cdot)$ and $D_{\text{emb}}(\cdot)$ denote learnable modality and day embedding lookups, respectively.
We denote the resulting decoder input embedding sequence as $F_{\mathrm{emb}} := (h_1,\ldots,h_L)$.

\subsubsection*{Multimodal Readmission-Risk Prediction Decoder}
\label{sec:model}

We model the admission trajectory by feeding the decoder input $\tilde{F}_{\mathrm{emb}}=[F_{\mathrm{emb}}, \texttt{[READMIT]}, \texttt{[SupCon]}]$ into a causal Transformer decoder (Fig.~\ref{fig:model}).
Masked self-attention enforces chronological information flow: each position aggregates information only from preceding tokens, so the appended summary tokens can attend to the entire admission stream while respecting its temporal order.

\paragraph{Training objective.}
We jointly optimize two objectives using the appended summary tokens in
$\tilde{F}_{\mathrm{emb}}$.
Let $\mathbf{z}_{\text{readmit}}$ and $\mathbf{z}_{\text{supcon}}$
denote the final hidden states of \texttt{[READMIT]} and
\texttt{[SupCon]}, respectively.
We map $\mathbf{z}_{\text{readmit}}$ to the admission-level
probability $\hat{y}$ using an MLP classification head, and map
$\mathbf{z}_{\text{supcon}}$ to a projected representation
$\tilde{\mathbf{z}}_{\text{supcon}}$ using a separate MLP projection
head.
This decouples the predictive and contrastive signals, and the model
is trained end-to-end with
\[
\mathcal{L}_{\text{total}} = \mathcal{L}_{\text{readmit}} +
\lambda\,\mathcal{L}_{\text{SupCon}},
\]
where $\lambda=0.3$ scales the contrastive contribution.

\textit{Readmission prediction.}
We supervise the admission-level prediction with binary cross-entropy:
\begin{equation}
\mathcal{L}_{\text{readmit}}
=
-\big[y\log\hat{y} + (1-y)\log(1-\hat{y})\big].
\end{equation}

\textit{Disease-specific supervised contrastive learning.}
We apply supervised contrastive loss to the projected representation $\tilde{\mathbf{z}}_{\text{supcon}}$ to impose diagnosis-supergroup-aware structure in the latent space and improve robustness to noisy readmission labels. We define positives using a coarse diagnostic label \texttt{SUPER\_GROUP}\footnote{We use diagnostic super-groups with the following categories: Circulatory (19.4\%), Endocrine \& Metabolic (13.1\%), Symptoms \& Signs (8.4\%), Respiratory (7.9\%), Trauma \& Poisoning (7.6\%), Digestive (7.2\%), Genitourinary (5.9\%), Factors \& Services (5.2\%), Blood \& Immune (5.2\%), Mental Disorders (4.9\%), Nervous System (4.5\%), Musculoskeletal \& Skin (4.5\%), Infectious (3.7\%), and Neoplasms (2.5\%).}; each admission is assigned a single \texttt{SUPER\_GROUP} as its most frequent coarse ICD-10 category (14 total).
For an anchor admission $i$ in a batch $\mathcal{B}$, positives $\mathcal{P}(i)$ are admissions that share both the same readmission label $y$ and the same \texttt{SUPER\_GROUP}.
To ensure positives exist for all anchors, each mini-batch includes at least two samples for every (\texttt{SUPER\_GROUP}, $y$) combination, while combinations are sampled proportionally to their prevalence.
Here, $\mathrm{sim}(\cdot,\cdot)$ denotes cosine similarity and $\tau=0.1$ is a temperature hyperparameter.
Let $\tilde{\mathbf{z}}_i$ denote the projection-head output of the \texttt{[SupCon]} token for admission $i$.
The loss is
\begin{equation}
\mathcal{L}_{\text{SupCon}}
=
\sum_{i \in \mathcal{B}}
\frac{-1}{|\mathcal{P}(i)|}
\sum_{p \in \mathcal{P}(i)}
\log
\frac{\exp(\mathrm{sim}(\tilde{\mathbf{z}}_i,\tilde{\mathbf{z}}_p)/\tau)}
{\sum_{a \in \mathcal{B}\setminus\{i\}} \exp(\mathrm{sim}(\tilde{\mathbf{z}}_i,\tilde{\mathbf{z}}_a)/\tau)}.
\end{equation}

\section{Experiments}
\subsection{Experimental Setup}
\paragraph{Dataset.}
We conducted experiments on MIMIC-IV~\citep{PhysioNet-mimiciv-3.1}
linked with MIMIC-CXR~\citep{PhysioNet-mimic-cxr-2.1.0}.
Following prior studies~\citep{must, stgnn}, we include adult admissions (age $\ge$ 18) with a length of stay $\ge$ 48 hours and at least two CXRs during the stay. This yields 13,821 admissions from 11,972 patients.  
We apply a stratified patient-level split (90\%/10\%), yielding
12{,}438 training and 1{,}383 test admissions with no patient overlap.
Daily EHR summaries are constructed by concatenating demographics,
ICD-10 diagnosis subgroups, abnormal laboratory results, and
medications grouped by therapeutic class, with duplicates removed.
For imaging, we restrict to frontal-view radiographs (AP/PA).

\paragraph{Baselines.}
We compare \modelname{} against two categories of baselines. 
For a comprehensive evaluation, we include the reported metrics of MuST~\citep{must} as an established literature reference. 
To ensure a strictly fair and identical-cohort comparison, we prioritize a direct baseline evaluation by re-implementing MM-STGNN~\citep{stgnn} on our unified MIMIC-IV cohort.


\textit{LVLMs} (Qwen3-VL~\citep{qwen}, MedGemma~\citep{medgemma}, Lingshu~\citep{lingshu}) serve as an upper-bound probe for general-purpose multimodal reasoning. Given their scale (4B--32B parameters) and broad medical pretraining, they represent a strong zero-shot alternative to task-specialized trajectory modeling, assessing whether purpose-built longitudinal architectures remain necessary in the foundation model era.\footnote{Full prompting details for all LVLM baselines are available at \github{}.}

\subsection{Performance Analysis and Clinical Robustness}
Table~\ref{tab:sota} compares \modelname{} against all baselines
across modality settings via modality masking without retraining.
In the multimodal (CXR--EHR) setting, \modelname{} achieves the best overall performance (AUC 0.814, F1 0.752), outperforming the primary task-matched baseline MM-STGNN by +0.014 in AUC and +0.184 in F1. These gains are especially meaningful under class imbalance, where
ACC can be inflated by majority-class predictions: MM-STGNN attains high ACC (0.857) but much lower F1 (0.568), whereas \modelname{} improves positive-case detection while maintaining
competitive ACC (0.744).
Compared to its BCE-only ablation, \modelname{} shows comparable
AUC (0.814 vs.\ 0.809) but substantially higher Rec(N-C)
(0.548 vs.\ 0.329), confirming that $\mathcal{L}_{\text{SupCon}}$
improves sensitivity to non-critical readmission cases under label noise
and class imbalance—precisely where clinical value is highest.

\begin{table}[h!]
\centering
\caption{\footnotesize \textbf{Performance comparison for 30-day readmission prediction.} All models predict the same 30-day post-discharge outcome; the day limit refers to the length of the \emph{input} in-hospital sequence, not the prediction horizon. Unless otherwise noted, models are trained on sequences of up to 10 hospital days; \modelname{} (Max 30-days) extends the input window to 30 days. For the CXR+EHR setting, we report outcome-stratified recall ($y{=}1$): \textit{Rec(C)} for critical cases and \textit{Rec(N-C)} for non critical cases. `Params' and `Dom.' denote trainable parameters and training domain (Gen.: General, Med.: Medical), respectively.}
\label{tab:sota}
\setlength{\tabcolsep}{3.5pt}
\renewcommand{\arraystretch}{1.05}
\resizebox{\textwidth}{!}{
\begin{tabular}{lll rrr @{\hspace{0.8em}} rrr @{\hspace{0.8em}} rrrrr}
  \toprule
  \multicolumn{3}{c}{} &
  \multicolumn{3}{c}{\textbf{CXR}} &
  \multicolumn{3}{c}{\textbf{EHR}} &
  \multicolumn{5}{c}{\textbf{CXR+EHR}} \\
  \cmidrule(lr){4-6} \cmidrule(lr){7-9} \cmidrule(lr){10-14}
  \textbf{Model} & \textbf{Params} & \textbf{Dom.} &
  \textbf{AUC} & \textbf{ACC} & \textbf{F1} &
  \textbf{AUC} & \textbf{ACC} & \textbf{F1} &
  \textbf{AUC} & \textbf{ACC} & \textbf{F1} &
  \textbf{Rec(C)} & \textbf{Rec(N-C)} \\
  \midrule
  \multicolumn{14}{l}{\textit{Readmission-specialized models}} \\
  \addlinespace[2pt]
  MuST\footnotemark[1] & --- & Med.
    & 0.721 & 0.744 & ---
    & \textbf{0.784} & \textbf{0.850} & ---
    & 0.799 & 0.859 & --- & --- & --- \\
  MM-STGNN & 3M & Med.
    & 0.769 & 0.757 & 0.477
    & 0.781 & 0.848 & 0.556
    & 0.800 & 0.857 & 0.568 & 0.714 & 0.214 \\
  \addlinespace[2pt]
  \midrule
  \multicolumn{14}{l}{\textit{Large Vision-Language Models (LVLMs)}} \\
  \addlinespace[2pt]
  Qwen3-VL & 32B & Gen.
    & 0.525 & 0.550 & 0.595
    & 0.609 & 0.500 & 0.667
    & 0.605 & 0.505 & 0.669 & \textbf{1.000} & 0.007 \\
  MedGemma-1.5v & 4B & Med.
    & 0.546 & 0.554 & 0.625
    & 0.517 & 0.496 & 0.659
    & 0.461 & 0.462 & 0.600 & 0.753 & 0.030 \\
  MedGemma & 27B & Med.
    & 0.557 & 0.518 & \textbf{0.670}
    & 0.541 & 0.514 & 0.673
    & 0.541 & 0.500 & 0.666 & \textbf{1.000} & 0.007 \\
  Lingshu & 32B & Med.
    & 0.517 & 0.507 & 0.635
    & 0.561 & 0.527 & \textbf{0.676}
    & 0.489 & 0.518 & 0.672 & \textbf{1.000} & 0.038 \\
  \addlinespace[2pt]
  \specialrule{1.2pt}{1pt}{1pt}
  \multicolumn{14}{l}{\textit{Ours}} \\
  \addlinespace[2pt]
  \modelname{} (BCE only) & 19M & Med.
    & 0.717 & 0.633 & 0.483
    & 0.673 & 0.676 & 0.633
    & 0.809 & 0.722 & 0.697 & 0.821 & 0.329 \\
  \rowcolor{yellow!20}
  \modelname{} & 19M & Med.
    & \textbf{0.787} & 0.711 & 0.656
    & 0.750 & 0.707 & 0.665
    & \textbf{0.814} & 0.744 & \textbf{0.752} & 0.911 & \textbf{0.548} \\
  \modelname{} (Max 30-days) & 19M & Med.
    & 0.778 & 0.689 & 0.629
    & 0.697 & 0.661 & 0.672
    & 0.793 & 0.726 & 0.736 & 0.917 & 0.506 \\
  \bottomrule
\end{tabular}
}
\end{table}
\footnotetext[1]{Due to the absence of publicly available source
code, performance metrics for MuST are cited from the original
publication. F1-scores were not available and are denoted by `--'.}

To separate critical deterioration-driven sensitivity from non-critical readmission detection, we report outcome-stratified recall for readmission-positive admissions ($y=1$). Specifically, we stratify the positive cohort into two subgroups based on clinical severity at discharge: \textit{Critical} cases (i.e., end-of-life) and \textit{Non-critical} readmission cases, denoted as Rec(C) and Rec(N-C), respectively. LVLM baselines show high Rec(C) but near-zero Rec(N-C) (0.007--0.038), indicating that their sensitivity is heavily concentrated on critical cases. In contrast, \modelname{} preserves strong Rec(C) (0.911) while substantially improving Rec(N-C) to 0.548. 
\modelname{} (Max 30-days) extends the temporal window to 30 days for longer-stay admissions that baselines cannot natively handle. Despite the longer and noisier sequences, it remains highly competitive (AUC 0.793, F1 0.736) with a markedly higher F1 than MM-STGNN (+0.168), demonstrating graceful scalability without architectural modification.

\subsubsection{Consistent Performance Across Clinical Categories}

\begin{table*}[t]
\centering
\caption{\footnotesize Category-stratified performance for 30-day
readmission prediction in the paired CXR--EHR setting. Categories
are grouped by the primary source of clinical evidence
(radiographic findings vs.\ structured EHR).}
\label{tab:clinical_categories}
\small
\setlength{\tabcolsep}{2.6pt}
\renewcommand{\arraystretch}{1.05}
\begin{adjustbox}{width=\textwidth}
\begin{tabular}{@{}l ccc ccc ccc ccc ccc@{}}
  \toprule
  \multirow{3}{*}{\textbf{Model}} &
  \multicolumn{9}{c}{\textbf{Categories with Radiographic Manifestations}} &
  \multicolumn{6}{c}{\textbf{EHR-Predominant Categories}} \\
  \cmidrule(lr){2-10} \cmidrule(lr){11-16}
  & \multicolumn{3}{c}{\textbf{Circulatory}}
  & \multicolumn{3}{c}{\textbf{Respiratory}}
  & \multicolumn{3}{c}{\textbf{Infectious}}
  & \multicolumn{3}{c}{\textbf{Endocrine \& Metabolic}}
  & \multicolumn{3}{c}{\textbf{Genitourinary}} \\
  \cmidrule(lr){2-4}\cmidrule(lr){5-7}\cmidrule(lr){8-10}
  \cmidrule(lr){11-13}\cmidrule(lr){14-16}
  & \textbf{AUC} & \textbf{ACC} & \textbf{F1}
  & \textbf{AUC} & \textbf{ACC} & \textbf{F1}
  & \textbf{AUC} & \textbf{ACC} & \textbf{F1}
  & \textbf{AUC} & \textbf{ACC} & \textbf{F1}
  & \textbf{AUC} & \textbf{ACC} & \textbf{F1} \\
  \midrule
  \multicolumn{16}{l}{\textit{Readmission-specialized models}} \\
  \addlinespace[2pt]
  MM-STGNN &
  0.798 & \textbf{0.766} & 0.499 &
  0.794 & \textbf{0.745} & 0.514 &
  \textbf{0.799} & \textbf{0.733} & 0.543 &
  0.791 & \textbf{0.763} & 0.499 &
  \textbf{0.812} & 0.737 & 0.511 \\
  \addlinespace[2pt]
  \midrule
  \multicolumn{16}{l}{\textit{Large Vision-Language Models (LVLMs)}} \\
  \addlinespace[2pt]
  Qwen3-VL &
  0.587 & 0.525 & 0.687 &
  0.591 & 0.588 & 0.739 &
  0.585 & 0.558 & 0.714 &
  0.596 & 0.523 & 0.687 &
  0.571 & 0.567 & 0.724 \\
  MedGemma-1.5v &
  0.472 & 0.483 & 0.616 &
  0.458 & 0.500 & 0.654 &
  0.424 & 0.470 & 0.613 &
  0.473 & 0.480 & 0.615 &
  0.456 & 0.500 & 0.639 \\
  MedGemma &
  0.518 & 0.520 & 0.684 &
  0.513 & 0.584 & 0.737 &
  0.561 & 0.553 & 0.710 &
  0.512 & 0.521 & 0.685 &
  0.485 & 0.571 & 0.725 \\
  Lingshu &
  0.467 & 0.530 & 0.686 &
  0.432 & 0.584 & 0.734 &
  0.481 & 0.567 & 0.717 &
  0.478 & 0.528 & 0.686 &
  0.468 & 0.574 & 0.725 \\
  \addlinespace[2pt]
  \specialrule{1.2pt}{1pt}{1pt}
  \multicolumn{16}{l}{\textit{Ours}} \\
  \addlinespace[2pt]
  \modelname{} (BCE only) &
  0.803 & 0.716 & 0.699 &
  0.791 & 0.701 & 0.714 &
  0.794 & 0.692 & 0.720 &
  \textbf{0.800} & 0.712 & 0.701 &
  0.787 & 0.688 & 0.710 \\
  \rowcolor{yellow!20}
  \modelname{} &
  \textbf{0.815} & 0.747 & \textbf{0.761} &
  \textbf{0.798} & 0.732 & \textbf{0.771} &
  0.781 & 0.724 & \textbf{0.770} &
  \textbf{0.800} & 0.735 & \textbf{0.756} &
  0.800 & \textbf{0.743} & \textbf{0.782} \\
  \modelname{} (Max 30-days) &
  0.795 & 0.733 & 0.750 &
  0.761 & 0.716 & 0.760 &
  0.780 & 0.722 & 0.769 &
  0.789 & 0.720 & 0.739 &
  0.780 & 0.725 & 0.768 \\
  \bottomrule
\end{tabular}
\end{adjustbox}
\end{table*}

Table~\ref{tab:clinical_categories} reports category-stratified results for 30-day readmission prediction in the paired CXR--EHR setting. The five diagnostic groups span conditions with strong radiographic correlates (Circulatory, Respiratory, Infectious) as well as EHR-predominant categories where risk-relevant evidence is primarily reflected in structured daily records (Endocrine \& Metabolic, Genitourinary). Across all groups, \modelname{} maintains consistently strong performance (F1: 0.756--0.782), indicating that it leverages multimodal evidence in a category-appropriate manner. In contrast, general-purpose LVLM baselines exhibit only moderate discrimination, with substantially lower AUC (roughly 0.42--0.60), suggesting limited stability for specialized clinical risk stratification. Notably, \modelname{} retains robust category-wise performance when extending the temporal window to 30 days: \modelname{} (Max 30-days) achieves F1 of 0.739--0.769 across groups, supporting its scalability to longer admission trajectories without sacrificing consistency.

\subsubsection{Interpretability and Clinical Trajectory Analysis}

\begin{figure}[t]
  \centering
\includegraphics[height=0.3\textheight]{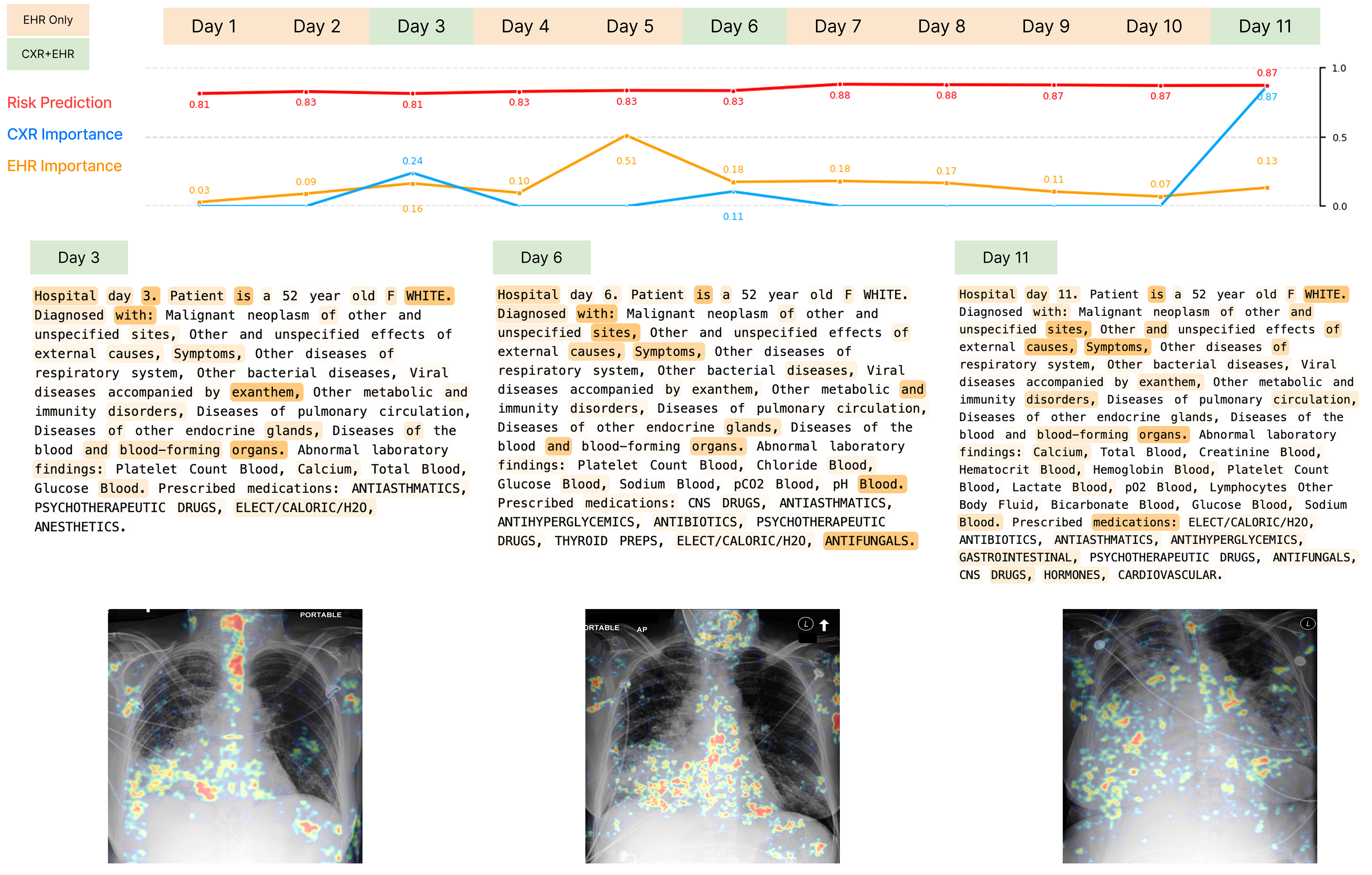}  
\caption{\footnotesize \textbf{Readmission prediction
in a deteriorating oncology admission.} CXRs were obtained only
on Hospital Days 3, 6, and 11 (shown with modality-specific
heatmaps). Risk rises from 0.81, peaks at 0.88 on Day 7, and remains at 0.87 through Day 11.}
  \label{fig:qualitative}
\end{figure}

Figure~\ref{fig:qualitative} illustrates \modelname{}'s
trajectory-level interpretability on a representative
deteriorating oncology admission. We define \textit{Critical Days}
as hospital days where token-level attributions peak and predicted
risk changes most markedly, providing temporally grounded signals
for real-time risk stratification. Heatmaps are derived by
back-projecting gradient-based importance scores from the
\texttt{[READMIT]} token through the decoder sequence to the
original EHR tokens and CXR patch embeddings.
CXRs were acquired only on Days 3, 6, and 11; modality-specific
heatmaps are shown at these time points. The predicted risk rises
from 0.81 to 0.87 by discharge, with EHR heatmaps emphasizing 
diagnosis and comorbidity tokens throughout, where Day 6 highlights 
anti-infective medications and Day 11 shifts toward 
abnormal-lab expressions. In parallel, CXR heatmaps show
lung-field activations expanding from Day 3 and peaking by
Day 11, together yielding temporally coherent cross-modal
explanations.

\section{Conclusion}
We presented \modelname{} (\textbf{M}ultimodal \textbf{r}eadmission-risk
prediction \textbf{Dec}oder), a framework that redefines readmission
prediction as a daily-scale causal trajectory modeling task.
By jointly modeling longitudinal EHR and intermittent CXR in a
single causal decoder, \modelname{} outperforms both graph-based
and large foundation model baselines, particularly in identifying
critical deterioration trajectories.
Disease-Specific Supervised Contrastive Learning further captures
inter-patient phenotypic structure without explicit graph
construction, and token-level interpretability provides
temporally grounded signals for real-time risk stratification.
Future work will explore extending the multimodal token stream
to additional clinical modalities such as ECG and clinical notes,
broadening the framework's applicability across diverse inpatient
settings.

\begin{credits}
\subsubsection{Acknowledgements}
This work was supported by the TIPS (Tech Incubator Program for Startup) Program of the Ministry of SMEs and Startups and Korea Startup Foundation under Grant No. RS-2025-25467010.

\subsubsection{Disclosure of Interests}
The authors have no competing interests to declare that are relevant to the content of this article.
\end{credits}

\bibliographystyle{splncs04}
\bibliography{ref}

@article{stgnn,
  title={Predicting 30-Day All-Cause Hospital Readmission Using Multimodal Spatiotemporal Graph Neural Networks},
  author={Tang, Siyi and Tariq, Amara and Dunnmon, Jared A and Sharma, Umesh and Elugunti, Praneetha and Rubin, Daniel L and Patel, Bhavik N and Banerjee, Imon},
  journal={IEEE Journal of Biomedical and Health Informatics},
  volume={27},
  number={4},
  pages={2071--2082},
  year={2023},
  publisher={IEEE}
}

@inproceedings{must,
  title={{MuST}: Multimodal Spatiotemporal Graph-Transformer for Hospital Readmission Prediction},
  author={Miao, Yan and Yu, Lequan},
  booktitle={International Conference on Medical Image Computing and Computer-Assisted Intervention},
  pages={276--285},
  year={2023},
  organization={Springer}
}

@inproceedings{diabetic_readmit,
  title={Hospital Readmission Risk Predictor for Diabetic Patients Using {XGBoost}},
  author={Reddy, Nareddy Koushik and Anirudh, Thummalapalli and Varshitha, Thondala and Kumar, Busetty Chethan and Kumar, NSSS Girish and Jabbar, MA},
  booktitle={2025 3rd World Conference on Communication \& Computing (WCONF)},
  pages={1--6},
  year={2025},
  organization={IEEE}
}

@inproceedings{alarcon2024prediction,
  title={Prediction of Readmission Risk in Hospital Patients Using Artificial Intelligence Techniques},
  author={Alarc{\'o}n, Iv{\'a}n Daniel Salazar and Santana-Vel{\'a}squez, Angelower and Duitama, M John Freddy and Salazar-S{\'a}nchez, Maria Bernarda and Hern{\'a}ndez-Arango, Alejandro},
  booktitle={2024 3rd International Congress of Biomedical Engineering and Bioengineering (CIIBBI)},
  pages={1--6},
  year={2024},
  organization={IEEE}
}

@article{nvidia,
  title={Health System-Scale Language Models Are All-Purpose Prediction Engines},
  author={Jiang, Lavender Yao and Liu, Xujin Chris and Nejatian, Nima Pour and Nasir-Moin, Mustafa and Wang, Duo and Abidin, Anas and Eaton, Kevin and Riina, Howard Antony and Laufer, Ilya and Punjabi, Paawan and others},
  journal={Nature},
  volume={619},
  number={7969},
  pages={357--362},
  year={2023},
  publisher={Nature Publishing Group UK London}
}

@article{social1,
  title={Social Determinants of Health and Hospital Readmission},
  author={Lax, Yonit and Martinez, Maria and Brown, Nicole M},
  journal={Pediatrics},
  volume={140},
  number={5},
  pages={e20171427},
  year={2017},
  publisher={American Academy of Pediatrics Elk Grove Village, IL, USA}
}

@article{social2,
  title={Social Determinants of Health and 30-Day Readmissions Among Adults Hospitalized for Heart Failure in the {REGARDS} Study},
  author={Sterling, Madeline R and Ringel, Joanna Bryan and Pinheiro, Laura C and Safford, Monika M and Levitan, Emily B and Phillips, Erica and Brown, Todd M and Nguyen, Oanh K and Goyal, Parag},
  journal={Circulation: Heart Failure},
  volume={15},
  number={1},
  pages={e008409},
  year={2022},
  publisher={Lippincott Williams \& Wilkins Hagerstown, MD}
}

@article{social3,
  title={Hospital Readmission and Social Risk Factors Identified from Physician Notes},
  author={Navathe, Amol S and Zhong, Feiran and Lei, Victor J and Chang, Frank Y and Sordo, Margarita and Topaz, Maxim and Navathe, Shamkant B and Rocha, Roberto A and Zhou, Li},
  journal={Health Services Research},
  volume={53},
  number={2},
  pages={1110--1136},
  year={2018},
  publisher={Wiley Online Library}
}

@incollection{readmission,
  title={Hospital Readmission Forecasting Using Artificial Intelligence},
  author={Subasi, Abdulhamit},
  booktitle={Applications of Artificial Intelligence in Healthcare and Biomedicine},
  pages={455--520},
  year={2024},
  publisher={Elsevier}
}

@article{readmission2,
  title={An Interpretable Machine Learning Approach for Predicting 30-Day Readmission After Stroke},
  author={Lv, Ji and Zhang, Mengmeng and Fu, Yujie and Chen, Mengshuang and Chen, Binjie and Xu, Zhiyuan and Yan, Xianliang and Hu, Shuqun and Zhao, Ningjun},
  journal={International Journal of Medical Informatics},
  volume={174},
  pages={105050},
  year={2023},
  publisher={Elsevier}
}

@article{okolie2025machine,
  title={Machine Learning Approaches for Predicting 30-Day Hospital Readmissions: Evidence from {Massachusetts} Healthcare Data},
  author={Okolie, Awele and Bello, A. and Ikhifa, M. O. and Ibiyeye, A. O. and Agbeso, D. O. and Alumona, P.},
  journal={World Journal of Advanced Research and Reviews},
  volume={28},
  number={1},
  pages={1--12},
  year={2025}
}

@misc{PhysioNet-mimiciv-3.1,
  title={{MIMIC-IV}},
  author={Johnson, Alistair and Bulgarelli, Lucas and Pollard, Tom and Gow, Brian and Moody, Benjamin and Horng, Steven and Celi, Leo Anthony and Mark, Roger},
  year={2024},
  howpublished={PhysioNet},
  note={Version 3.1, DOI: 10.13026/kpb9-mt58}
}

@misc{PhysioNet-mimic-cxr-2.1.0,
  title={{MIMIC-CXR} Database},
  author={Johnson, Alistair and Pollard, Tom and Mark, Roger and Berkowitz, Seth and Horng, Steven},
  year={2024},
  howpublished={PhysioNet},
  note={Version 2.1.0, DOI: 10.13026/4jqj-jw95}
}

@article{eva-x,
  title={{EVA-X}: A Foundation Model for General Chest {X}-Ray Analysis with Self-Supervised Learning},
  author={Yao, Jingfeng and Wang, Xinggang and Song, Yuehao and Zhao, Huangxuan and Ma, Jun and Chen, Yajie and Liu, Wenyu and Wang, Bo},
  journal={npj Digital Medicine},
  volume={8},
  number={1},
  pages={678},
  year={2025},
  publisher={Nature Publishing Group UK London}
}

@misc{modernbert,
  title={{BioClinical ModernBERT}: A State-of-the-Art Long-Context Encoder for Biomedical and Clinical {NLP}},
  author={Sounack, Thomas and Davis, Joshua and Durieux, Brigitte and Chaffin, Antoine and Pollard, Tom J and Lehman, Eric and Johnson, Alistair EW and McDermott, Matthew and Naumann, Tristan and Lindvall, Charlotta},
  year={2025},
  eprint={2506.10896},
  archivePrefix={arXiv},
  primaryClass={cs.CL}
}

@article{jiang2019readmission,
  title={Readmission Risk Trajectories for Patients with Heart Failure Using a Dynamic Prediction Approach: Retrospective Study},
  author={Jiang, Wei and Siddiqui, Sauleh and Barnes, Sean and Barouch, Lili A and Korley, Frederick and Martinez, Diego A and Toerper, Matthew and Cabral, Stephanie and Hamrock, Eric and Levin, Scott},
  journal={JMIR Medical Informatics},
  volume={7},
  number={4},
  pages={e14756},
  year={2019}
}

@article{davis2024survival,
  title={Survival Models and Longitudinal Medical Events for Hospital Readmission Forecasting},
  author={Davis, Sacha and Greiner, Russell},
  journal={BMC Health Services Research},
  volume={24},
  number={1},
  pages={1394},
  year={2024},
  publisher={Springer}
}

@misc{qwen,
  title={{Qwen3} Technical Report}, 
  author={{Qwen Team}},
  year={2025},
  eprint={2505.09388},
  archivePrefix={arXiv},
  primaryClass={cs.CL}
}

@misc{medgemma,
  title={{MedGemma} Technical Report},
  author={Sellergren, Andrew and Kazemzadeh, Sahar and Jaroensri, Tiam and Kiraly, Atilla and Traverse, Madeleine and Kohlberger, Timo and Xu, Shawn and Jamil, Fayaz and Hughes, C{\'\i}an and Lau, Charles and others},
  year={2025},
  eprint={2507.05201},
  archivePrefix={arXiv},
  primaryClass={cs.CL}
}

@misc{lingshu,
  title={{Lingshu}: A Generalist Foundation Model for Unified Multimodal Medical Understanding and Reasoning},
  author={Xu, Weiwen and Chan, Hou Pong and Li, Long and Aljunied, Mahani and Yuan, Ruifeng and Wang, Jianyu and Xiao, Chenghao and Chen, Guizhen and Liu, Chaoqun and Li, Zhaodonghui and others},
  year={2025},
  eprint={2506.07044},
  archivePrefix={arXiv},
  primaryClass={cs.CL}
}

\end{document}